\documentclass[10pt]{elsarticle}
\usepackage{lineno}
\usepackage{booktabs}
\usepackage{multirow}
\usepackage[hang,small,bf]{caption}
\usepackage[subrefformat=parens]{subcaption}
\usepackage{url}
\usepackage[top=20truemm,bottom=20truemm,left=25truemm,right=25truemm]{geometry}
\usepackage{color}
\usepackage{bm}
\usepackage{xr}
\usepackage{amsmath, amssymb}
\usepackage{type1cm}
\usepackage{comment}
\usepackage{seqsplit}
\usepackage{ulem}
\title{GraphIX: Graph-based In silico XAI(explainable artificial intelligence) for drug repositioning from biopharmaceutical network}
\author{Atsuko Takagi, Mayumi Kamada, Eri Hamatani, Ryosuke Kojima, Yasushi Okuno}

\makeatletter
\newcommand*{\addFileDependency}[1]{
  \typeout{(#1)}
  \@addtofilelist{#1}
  \IfFileExists{#1}{}{\typeout{No file #1.}}
}
\makeatother

\newcommand*{\myexternaldocument}[1]{
    \externaldocument{#1}
    \addFileDependency{#1.tex}
    \addFileDependency{#1.aux}
}

\myexternaldocument{anc/supplement}

\begin{document}
\date{}
\begin{abstract}
Drug repositioning holds great promise because it can reduce the time and cost of new drug development. While drug repositioning can omit various R\&D processes, confirming pharmacological effects on biomolecules is essential for application to new diseases. Biomedical explainability in a disease-drug association prediction model is able to support appropriate insights in subsequent in-depth studies. However, the validity of the XAI methodology is still under debate, and the effectiveness of XAI in drug repositioning prediction applications remains unclear.\\

In this study, we propose GraphIX, an explainable drug repositioning framework using biological networks, and quantitatively evaluate its explainability. GraphIX first learns the network weights and node features using a graph neural network from known drug indication and knowledge graph that consists of three types of nodes (but not given node type information): disease, drug, and protein. Analysis of the post-learning features showed that node types that were not known to the model beforehand are distinguished through the learning process based on the graph structure. From the learned weights and features, GraphIX then predicts the disease-drug association and calculates the contribution values of the nodes located in the neighborhood of the predicted disease and drug. We hypothesized that the neighboring protein node to which the model gave a high contribution is also important in understanding the actual pharmacological effects. Quantitative evaluation of the validity of protein nodes’ contribution using a real-world database showed that the high contribution proteins shown by GraphIX are reasonable as a mechanism of drug action. GraphIX is a drug repositioning framework for evidence-based drug discovery that can present to users new disease-drug associations and identify the protein important for understanding its pharmacological effects from a large and complex knowledge base.\\
\end{abstract}
\maketitle

\section{Introduction} 
Drug repositioning, which converts a known drug into a treatment for a new disease, can reduce costs because the safety verification process can be skipped. Specifically, the estimation in [1] suggests that drug development time can be cut to six years and an average of \$300 million. Drug repositioning holds great promise because it can enable us to provide drugs at a low cost in a short period. \\

Computational methods are well suited for drug repositioning because it requires the identification of a new association with a disease for a vast number of known drugs. Since machine learning can use known data and knowledge graphs can directly represent the relationships between various entities, methods employing these two have been developed\cite{Wang2013}\cite{Chen2020}\cite{Wang2020}. For example, Wang et al. proposed PreDR, which learns associations from known disease-drug pair features using support vector machine\cite{Wang2013}. BiFusion is a framework that uses Graph Attention Network to learn disease-drug bipartite graph and protein network\cite{Wang2020}. All of these machine learning-based methods successfully achieved high association prediction accuracy.\\

On the other hand, since evidence-based drug discovery has become mainstream, confirming pharmacological effects on biomolecules is essential when applying them to new diseases in drug repositioning. Therefore, it is desirable for the prediction model to be able to provide biomedical interpretations, and the explainability of the prediction results may support appropriate insights in subsequent in-depth studies. In recent years, several drug repositioning methods have also been developed that can present explainability, i.e., which entities contribute to the predicted results\cite{Guney2016}\cite{Taie2021}\cite{He2022}. For example, Guney et al. introduced that the disease-drug associations can be statistically explained by the proximity; the shortest path length from the drug targets to the diseased proteins on the molecular network\cite{Guney2016}. Elsewhere, EDEN used Attention to present an explanation path connecting the disease to the drug in the Knowledge Graph of Biomedical Entities\cite{He2022}.\\

However, these approaches were either unclear as to which of the numerous candidate proteins were truly important, or they only provided examples of relatively high contributing explanation paths that could be validated in the literature. In other words, although they focus on explainability, they remain less clear about the validity of their explainability in drug repositioning applications. Since the explainability of the XAI method is still under debate and the inadequacy of testing only anecdotal case evidence is pointed out, validation that includes quantitative evaluation has become important and the percentage of papers providing quantitative evaluations of XAI methods is increasing every year\cite{Rudin2019}\cite{Nauta2022}. In short, in drug repositioning models, it is very important not only focusing on association prediction accuracy and explainability but also on how reasonable explainability the model has as the next step.\\

In this study, we introduce GraphIX(short for a \uline{Graph}-based, \uline{I}n silico, \uline{X}AI drug repositioning framework). GraphIX predicts drug-disease associations by representing the associations among drugs, diseases, and proteins as a large knowledge graph and learning the entire graph structure using a graph neural network(GNN). Then, it calculates the contribution values of the proteins located near the predicted disease and drug on the knowledge graph using Integrated Gradients(IG).\\
The main contributions of this work include:
\begin{itemize}
\item The node types, that were not given to the model in the input were found to be distinguished in the learning process based on the graph structure.
\item Quantitative evaluation using a real-world database demonstrated the validity of this model’s explainability in drug repositioning applications.
\item Since the predicted target node types are interchangeable, the method is applicable not only to drug repositioning but also to target repositioning and disease-related protein discovery.
\end{itemize}

\section{Materials and methods}
\subsection{Model framework}
\begin{figure}[h]
\centering
\includegraphics[width=16cm]{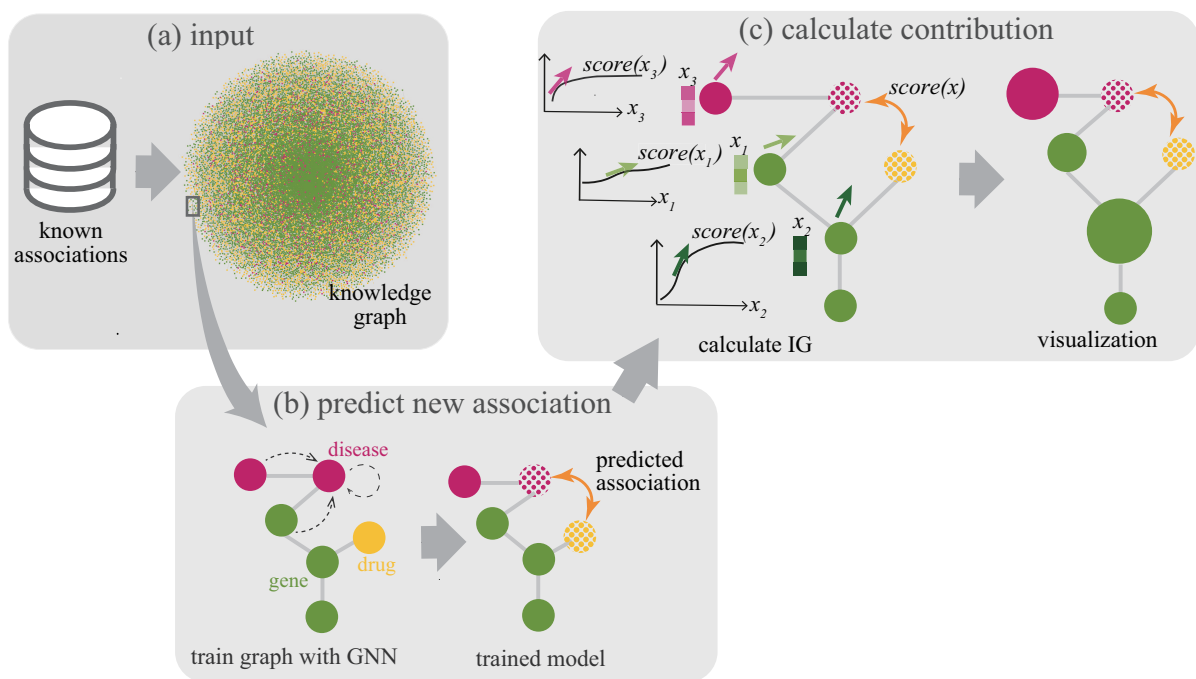}
\caption{Overview of GraphIX (Prediction of disease-drug association).}
\label{fig:Overview}
\end{figure}

An overview of GraphIX is shown in Fig.\ref{fig:Overview}. The framework consists of mainly two parts: the association prediction and the contribution calculation. 
In the association prediction part, the target association (disease-drug in drug repositioning) is predicted. That is, we trained a model on the knowledge graph using GNN to predict new associations from known ones. 
The calculation of contribution interprets the drug repositioning prediction above. To be concrete, we apply Integrated Gradients (IG)\cite{Sundararajan2017} to the trained model to calculate the contribution of disease-drug prediction for surrounding nodes and visualize them to present genes that are important for prediction.\\

\subsection{Construction of knowledge graph}
GraphIX accepts a knowledge graph of the associations of disease-drug, drug-gene, gene-gene, disease-gene, and disease-disease as input data. Here, the gene node represents the protein encoded by the gene. 
The disease-drug and drug-gene associations were obtained from the dataset compiled by Guney et al.\cite{Guney2017}. The disease-drug associations were collected from drug indication information in public databases such as KEGG and Metab2MeSH, and the drug-gene associations were constructed from drug-target information on FDA-approved drugs in the DrugBank database.
For gene-gene and disease-gene associations, we adopted datasets edited by Menche et al.\cite{Menche2015}. The gene-gene associations integrate various types of protein interactions from public databases such as TRANSFAC and  IntAct. The disease-gene associations were collected from OMIM and GWAS catalog. In this dataset, the associations were expanded from a gene-related specific disease upward to the most general ones along MeSH Hierarchy. Thus, In this study, to make the graph connections simple, we converted the disease names to MeSH Tree numbers and removed the trivial edges upward of specific diseases. Furthermore, to represent the relationship between each specific disease node, we merged the MeSH disease tree structure into the knowledge graph as disease-disease associations (See Supplementary Fig.\ref{fig:KGwithMeSH} for the details).\\

Depending on the association to be predicted, the dataset on which the knowledge graph is constructed is changed. For example, in predicting the disease-drug associations as a drug repositioning, we targeted four associations (drug-gene, gene-gene, disease-gene, disease-disease), excluding the disease and the drug, and merged those associations. Then, the largest network of each association was extracted from the merged associations to create the final knowledge graph.\\

\subsection{Model training for association prediction}
For the association prediction, graph convolutional network\cite{Kipf2016}\cite{Schlichtkrull2018} was employed to learn the structure of the biopharmaceutical knowledge graph generated above. Given an undirected graph with total $N$ nodes consisting of $R$ relation types, including labeled $P$ target association pairs to be predicted (e.g., disease-drug). The negative pairs $P'$ are randomly selected from the same type of association pairs to equal the number of positive examples. The model consists of an embedding layer and a convolution layer. In the embedding layer, each node feature is initially represented by a vector $\bm{x} \in{\mathbb{R}^{C}}$ with random values and updated by learning. In the \textit{l}-th convolution layer, node features are updated by the following propagation rule:
\begin{equation}
    \label{forward}
    \bm{X}^{(l+1)} = \sigma (\sum_{r\in{R}}\bm{A}_{r}\bm{X}^{(l)}\bm{W}_r^{(l)})
\end{equation}
Here, $\bm{X}^{(l)}\in{\mathbb{R}^{N×C}}$ is the matrix of node feature vectors, $\sigma(\cdot)$ is the Tanh activation function, $\bm{A_r}$ is the adjacency matrix under relation $r\in{R}$, $\bm{W^{(l)}_{r}}\in{\mathbb{R}^{C×D}}$ is a weight matrix with $D$ being the dimensionality after convolution. After updating $\bm{X}$ by forward propagation, we defined an association score $f$ between two nodes $(i, j)$ using dot product:
\begin{equation}
    \label{score}
    f(\bm{x}_i, \bm{x}_j) = \bm{x}_i \cdot \bm{x}_j
\end{equation} 
To train the model, we employed a pairwise ranking loss as the objective function:
\begin{equation}
    \label{loss}
    \mathcal{L} = -\ln \left\{\mu\left(\sum_{(i,j)\in{P}}\sum_{(i',j')\in{P'}}f(\bm{x}_i, \bm{x}_j) - f(\bm{x}_i', \bm{x}_j')\right) + \epsilon\right\}
\end{equation}
where $\mu(\cdot)$ represents the Sigmoid function and $\epsilon = 1.0 \times 10^{-10}$. In this study, we set the number of convolution layers is set to 1, and the dimension of node feature vectors $C$ and $D$ are set to 64. $R$ is a total of 5 with four types of associations and self-connections.\\

\subsection{Performance evaluation}
\subsubsection{Comparison with machine learning method}
To validate the prediction accuracy of the proposed approach, a comparative validation with existing machine learning-based methods was conducted. Three well-known supervised learning methods for network-based prediction, BiFusion\cite{Wang2020}, TransE\cite{Bordes2013}, and DistMult\cite{Yang2014}, were used as comparators. BiFusion\cite{Wang2020} is an encoder-decoder model that uses message passing on a graph. The encoder propagates node features by applying graph attention convolution to bipartite graphs for the disease-protein and drug-protein and the global graph representing PPI(Protein-Protein Interaction). The decoder is a multilayer perceptron that learns associations from disease and drug node features. 
Knowledge in multi-relational data is represented as a collection of triplets $(head,\ relation,\ tail)$, where head and tail are concrete or abstract entities and relation is a predicate representing the relationship between the two. The vector representation of $(head,\ relation,\ tail)$ is written as $(\bm{h}, \bm{r}, \bm{t})$, respectively. TransE\cite{Bordes2013} represents the relationship between head and tail as an additive interaction, and its score function(L1 formulation) is defined by $f(h,r,t) = \parallel \bm{h} + \bm{r} - \bm{t} \parallel_{1}$.
DistMult\cite{Yang2014} represents the relationship as a multiplicative interaction, and its score function is defined by $f(h,r,t) = \langle \bm{h}, \bm{r}, \bm{t} \rangle$, where $\langle \cdot \rangle$ denotes the generalized dot product. Note that for TransE and DistMult, the single relation vector $\bm{r}$ was used because of evaluating the performance for predicting disease-drug association. 
We used distributed representation for features of diseases, drugs, and proteins. 
The dataset on GitHub provided by Guney et al.\cite{Guney2017} was used for the training dataset. Disease names were converted to MeSH tree numbers based on the steps described in the previous section. The relevance data required for each method was also constructed using the same dataset as GraphIX. 
The 5-fold cross-validation was conducted, and the means of Area Under the Receiver Operatorating Characteristic curve(ROC-AUC) and Area Under the Precision-Recall curve(PR-AUC) were used to evaluate prediction performance. \\
By rotating node types, the association between disease-gene and drug-gene can be predicted in the same way. Therefore, we compared the means of ROC-AUC and PR-AUC of the method with machine learning methods for associations other than disease-drug. The BiFusion method requires drug-drug association data to predict disease-gene association. However, since this data could not be obtained from the target data, it was excluded from the evaluation. \\

\subsubsection{Comparison with non-machine learning method}
Accuracy comparisons were also conducted with the network-based proximity method\cite{Guney2016}, the non-machine learning method of prediction results. The network-based proximity method is the statistical approach that measures relative proximity, defined by the shortest path length between the disease proteins and the drug targets in the PPI network. 
We employed the disease-drug associations used for evaluation in the paper by Guney et al.\cite{Guney2016} as performance evaluation data.
To be precise, for the drug-disease associations between 78 diseases and 238 drugs in the paper, 402 known pairs were selected as positive examples and an equal number of randomly selected pairs from the remaining unknown pairs as negative examples.
In GraphIX prediction, since GraphIX converts disease names to hierarchal MeSH tree numbers, 78 disease names were expanded to 147 tree numbers, resulting in 402 positive disease-drug pairs were to 856 pairs. After predicting association scores for each edge, we took the average scores for each disease and used them for inference performance evaluation.
All remaining disease-drug associations published in the network-based proximity GitHub\cite{Guney2017} were used to train the model, with the exception of the inference set.
We drew the receiver operating characteristic (ROC) curve using the proximity value provided in Supplementary Data1 of the paper for network-based proximity and the association scores on the inference set for GraphIX.\\

\subsection{Calculation of contributions}
In the contribution calculation part, Integrated Gradients (IG) were used to identify the entities (nodes) that contributed to the association prediction. The IG of node $i$ for an association score is defined as:
\begin{equation}
    \label{IG}
   IG(\bm{x_{i}}) = \|\bm{x_i}\sum_{k=1}^{m}\Delta F(\frac{k}{m}\bm{x_i})\frac{1}{m}\|_{2}
\end{equation}
Here, $\bm{x_i}$ represents the feature vector of node $i$, $\|\:\|_{2}$ is the L2 norm, and $F(\bm{x_i})$ is the function of $\bm{x_i}$ for the predicted score. IG is defined as the sum of the gradients of $F(\bm{x_i})$ when the features of node $i$ are changed from 0 to the value after training in $m$ steps. $m$ is set to 30 in this study. The L2 norm converts the IG vector, which has the same dimensionality as the features, into the magnitude of a vector for each node. Let k-hop denote that the two vertices of the graph are connected by the k shortest path edges. In the k-layer convolution model, the feature vector values of the nodes within k-hop are propagated, and the association score for each edge is calculated. Therefore, this time, IG is calculated for nodes within 1-hop of the nodes forming the edge to be predicted.\\

\subsection{Quantitative verification of explainability}
The required explainability of the predictive model is that the entities used for the prediction can be identified and that the entities are reasonable for the target being predicted.
The target protein of a molecularly targeted drug is directly related to the relationship between the targeted drug and the disease. Therefore, in this study, we quantitatively evaluated the explainability of drug repositioning prediction utilizing available cancer type - molecularly targeted drug - target protein sets obtained from the cancer knowledge base OncoKB\cite{Chakravarty2017}. Specifically, we evaluated whether the protein node with the highest IG for the predicted association score between cancer type (disease node) and molecularly targeted drug (drug node) matched the target protein in OncoKB.
First, we excluded data that did not exist in the knowledge graph, such as fusion genes, from the OncoKB cancer type-molecular target-drug-target protein set. Then, we extracted sets of cancer types and molecularly targeted drugs predicted to be positive in the knowledge graph. 
The one-layer GNN model in this study allows IG values to be calculated for nodes within 1-hop of the predicted disease-drug edge. Therefore, only the sets where the target protein node was 1-hop from the disease-drug edge were included, resulting in 21 sets of disease-drug-target proteins for evaluation. 
For each of these 21 disease-drugs associations, the IG values of the surrounding 1-hop nodes were calculated and ordered in descending order focusing on the protein nodes. Then we evaluated the proteins with the highest IG values based on how well they matched the known target proteins.

\subsection{Case studies for novel association prediction}
We conducted a literature-based case study to further test the effectiveness of GraphIX's novel association predictions and explainability. For associations not included in the training data, we first excluded associations involving drugs classified as dietary supplements in DrugBank. For associations involving diseases, because associations with synonymous diseases (e.g., Nervous System Neoplasms and Neuroblastoma are regarded as synonymous diseases) may be included in the training data, we excluded synonymous associations by defining synonyms using a MeSH tree. Specifically, diseases that shared the first level of the MeSH tree hierarchy were considered synonyms, and if the corresponding disease and drug associations were included in the training data, they were considered synonymous associations and excluded from the validation. For the case study of the disease-drug prediction, the scores for novel associations were sorted in descending order by disease, and conducted a literature review for the top 10 associations and the protein node with the highest IG for the top association. The case studies for other association predictions were conducted similarly. For the disease-gene predictions, so we excluded in this study because no literature was found to support the results. CytoScape\cite{Shannon2003} was used to visualize the explainability using IG of the novel association predictions.\\

\section{Results and Discussion} 
\subsection{Performance evaluation}

\begingroup
\renewcommand{\arraystretch}{1.4}
\begin{table}[h]
\begin{minipage}{16cm}
\centering
\begin{tabular}{ccccccccc}
\toprule
\multirow{2}{*}{\begin{tabular}{c}\# positive \\ disease-drug\end{tabular}} && \multicolumn{4}{c}{\# edges} & \multicolumn{3}{c}{\# nodes} \\
\cline{2-5} \cline{7-9}
 & disease-disease & disease-gene & gene-gene & gene-drug & & disease & drug & gene \\
\midrule
35260 & 4411 & 16734 & 141217 & 11913 && 4412 & 4925 & 15164 \\
\bottomrule
\end{tabular}
\subcaption{Disease-drug prediction.}
\end{minipage}

\begin{minipage}{16cm}
\vspace{0.5cm}
\end{minipage}

\begin{minipage}{16cm}
\centering
\begin{tabular}{ccccccccc}
\toprule
\multirow{2}{*}{\begin{tabular}{c}\# positive \\ drug-gene\end{tabular}} & \multicolumn{4}{c}{\# edges} && \multicolumn{3}{c}{\# nodes} \\
\cline{2-5} \cline{7-9}
 & disease-disease & disease-gene & gene-gene & disease-drug & & disease & drug & gene \\
\midrule
12421 & 4411 & 16734 & 141169 & 35260 && 4412 & 1420 & 14088 \\
\bottomrule
\end{tabular}
\subcaption{Drug-gene prediction.}
\end{minipage}

\begin{minipage}{16cm}
\vspace{0.5cm}
\end{minipage}

\begin{minipage}{16cm}
\centering
\begin{tabular}{ccccccccc}
\toprule
\multirow{2}{*}{\begin{tabular}{c}\# positive \\ disease-gene\end{tabular}} & \multicolumn{4}{c}{\# edges} && \multicolumn{3}{c}{\# nodes} \\
\cline{2-5} \cline{7-9}
 & disease-disease & gene-gene & disease-drug & drug-gene & & disease & drug & gene \\
\midrule
16734 & 4411 & 141202 & 35260 & 11984 && 4412 & 5130 & 14488 \\
\bottomrule
\end{tabular}
\subcaption{Disease-gene prediction.}
\end{minipage}
\caption{Network components.}
\label{table:Components}
\end{table}
\endgroup

\begin{figure}[h]
\centering
\includegraphics[width=10cm]{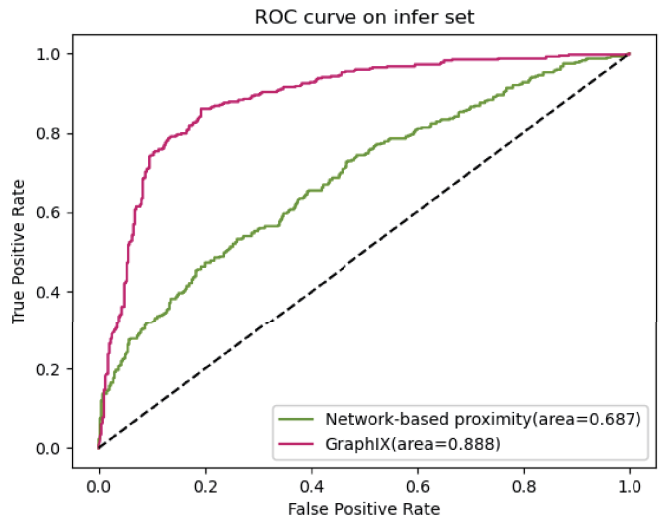}
\caption{ Receiver operating characteristic(ROC) curves of GraphIX and network-based proximity on infer dataset.}
\label{fig:infer}
\end{figure}

\begingroup
\renewcommand{\arraystretch}{1.4}
\begin{table}[h]
\centering
\begin{tabular}{lcccccccc}
\toprule
\multirow{2}{*}{Method} & \multicolumn{2}{c}{Disease-drug} && \multicolumn{2}{c}{Gene-drug} && \multicolumn{2}{c}{Disease-gene} \\
\cline{2-3} \cline{5-6}  \cline{8-9}
 & ROC-AUC & PR-AUC && ROC-AUC & PR-AUC && ROC-AUC & PR-AUC \\
\midrule
BiFusion & 0.976±0.001 & 0.967±0.002 && 0.966±0.004 & 0.969±0.005 && - & - \\
TransE   & 0.978±0.002 & 0.971±0.003 && 0.955±0.006 & 0.960±0.005 && 0.957±0.001 & 0.953±0.002 \\
DistMult & 0.987±0.002 & 0.987±0.002 && 0.924±0.008 & 0.954±0.005 && 0.958±0.002 & 0.975±0.001 \\
GraphIX  & 0.992±0.003 & 0.990±0.004 && 0.935±0.011 & 0.953±0.006 && 0.952±0.003 & 0.964±0.005 \\
\bottomrule
\end{tabular}
\caption{The mean of ROC-AUC and PR-AUC for 5-fold cross-validation on each prediction task.\\ For Gene-drug prediction, BiFusion was not evaluated since the drug-drug relationships required to create the model could not be obtained from the target data.}
\label{table:Accuracy-focused}
\end{table}
\endgroup

The knowledge graphs were constructed for each association type to be predicted and used to learn and predict. The components of each graph are shown in Table\ref{table:Components}. First, we evaluated the prediction performance of GraphIX and compared it with these of conventional machine learning-based methods. BiFusion and GraphIX learn graph structure information, and TransE and DistMult learn vector similarity through transition functions. Table\ref{table:Accuracy-focused} shows the mean ROC-AUC and PR-AUC for disease-drug, gene-drug, and disease-gene predictions. The results show that all the learning frameworks show high performance, and GraphIX can make predictions in all types of tasks with accuracy comparable to conventional high-accuracy machine learning methods. Next, we compared GraphIX with the network-based proximity method\cite{Guney2016}. Both are graph-based models with explainability, but the network-based proximity method is a non-machine learning framework, while GraphIX uses a supervised learning framework. As Fig. \ref{fig:infer} clearly shows, GraphIX significantly outperformed the network-based proximity method. These results indicate the effectiveness of using a learning framework for prediction and that GraphIX has accuracy comparable to state-of-the-art high-performance machine learning methods.\\

\subsection{Quantitative verification of explainability}
\begin{figure}[h]
\centering
\includegraphics[width=18cm]{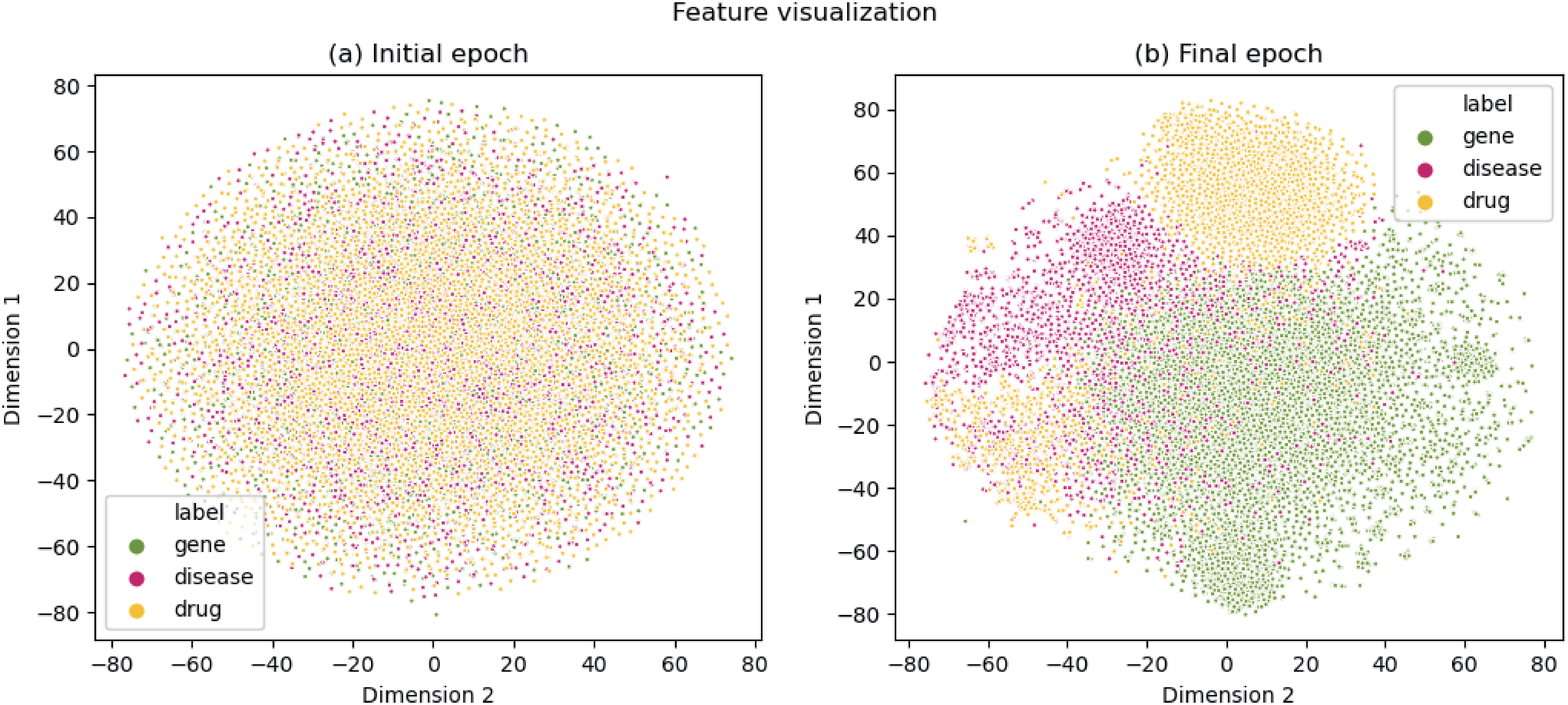}
\captionsetup{justification=raggedright,singlelinecheck=false}
\caption{Feature visualization via tSNE of all nodes in the graph at (a) initial epoch and (b) final epoch. The first fold, which is employed in inference and IG calculation is used.}
\label{fig:FeatureHeatmap}
\end{figure}

\begin{figure}[h]
\centering
\includegraphics[width=12cm]{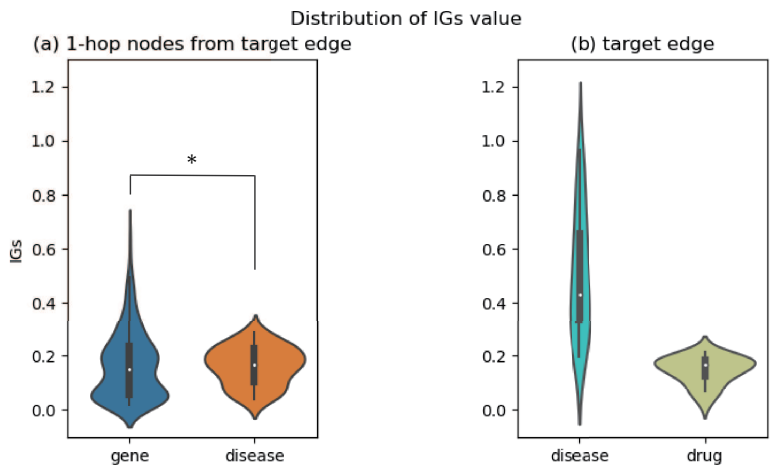}
\caption{Value of IG on (a)1-hop nodes from target edge and (b)target edge, represented via violin plots. Nonparametric statistics Anderson-Darling test is used for significant difference analysis. * indicates p-value \textless 0.05 for the test between 1-hop disease and gene node's IG distributions.}
\label{fig:IGDistribution}
\end{figure}

\begingroup
\renewcommand{\arraystretch}{1.2}
\begin{table}[!hbp]
\centering
\begin{tabular}{cccccc}
\toprule
Disease & Drug & Target protein && \# \multirow{2}{*}{\begin{tabular}{c}Candidate \\proteins\end{tabular}} & \multirow{2}{*}{\begin{tabular}{c}Ranking of target proteins \\within candidate proteins\end{tabular}} \\
 & & && & \\
\midrule
\multirow{1}{*}{\begin{tabular}{c}Acute Myeloid \\ Leukemia\end{tabular}} 
                            & Sorafenib & FLT3 && 38 & 4 \\
                            &           &      &&    &   \\ \\
\multirow{4}{*}{\begin{tabular}{c}Chronic\\ Myelogenous\\ Leukemia\end{tabular}} 
                            & Dasatinib & ABL1 && 15 & 1 \\
                            & Nilotinib & ABL1 && 7  & 1 \\
                            & Bosutinib & ABL1 && 14 & 1 \\
                            & Ponatinib & ABL1 && 19 & 2 \\ \\
\multirow{7}{*}{\begin{tabular}{c}Gastrointestinal\\ Stromal Tumor\end{tabular}} 
                            & Sorafenib   & KIT        && 11 & 1   \\
                            & Dasatinib   & PDGFRA     && 14 & 12  \\
                            & Nilotinib   & KIT        && 6  &  1  \\
                            & Regorafenib & KIT,PDGFRA && 21 & 1,4 \\
                            & Imatinib    & KIT,PDGFRA && 10 & 1,2 \\
                            & Pazopanib   & KIT        && 13 & 1   \\ 
                            & Sunitinib   & KIT,PDGFRA && 11 & 1,3 \\ \\
\multirow{1}{*}{Glioma} & Lapatinib & EGFR && 10 & 1 \\ \\
\multirow{4}{*}{Lymphoma} & Nilotinib & ABL1 && 3 & 2  \\
                          & Dasatinib & ABL1 && 11 & 2 \\ \\
\multirow{2}{*}{Melanoma} & Imatinib    & KIT  && 25 & 1 \\ 
                          & Vemurafenib & BRAF && 19 & 1 \\ \\
\multirow{2}{*}{\begin{tabular}{c}Non-Small Cell\\ Lung Cancer\end{tabular}}  
                            & Erlotinib & EGFR && 6 & 1 \\
                            & Gefitinib & EGFR && 5 & 1 \\ \\
\multirow{2}{*}{\begin{tabular}{c}Renal Cell\\ Carcinoma\end{tabular}} 
                            & Everolimus   & MTOR && 22 & 1 \\
                            & Temsirolimus & MTOR && 22 & 1 \\
\bottomrule
                            &             &      &&    & total accuracy = 16/21(\textcolor{red}{76\%})
\end{tabular}
\caption{The quantitative verification of explainability. Column 1-3 show cancer type, molecularly targeted drug, target protein in OncoKB evaluation set, respectively. Column 4 and 5 give the number of candidate protein nodes used by the model for prediction, the ranking of the IG value of the target protein among the candidate proteins, respectively. }
\label{table:OncoKB}
\end{table}
\endgroup

First, to test whether the model distinguishes between node types, we examined whether there is a difference in distribution with respect to node features and IGs after training. Regarding features, we used tSNE\cite{Maaten2008} to project to two dimensions for pre and post-training values for all nodes of the model used for quantitative verification of explainability. Fig. \ref{fig:FeatureHeatmap}(a) and Fig. \ref{fig:FeatureHeatmap}(b) show tSNE plots of features before and after training, respectively. Before training, the features were uniformly assigned random values for all node types, but after training, they were roughly clustered according to node types. Regarding IG, for 21 disease-drug-target protein evaluation sets extracted from the cancer knowledge database OncoKB\cite{Chakravarty2017}, we calculated IG values for disease and gene nodes at 1-hop from the predicted target(disease-drug) using a 1-layer GNN model. Regarding IG, for 21 disease-drug-target protein evaluation sets extracted from the cancer knowledge database OncoKB, we calculated IG values for disease and gene nodes at 1-hop from the predicted target(disease-drug) using a 1-layer GNN model. There is a significant difference in the distribution of IG values for disease and gene nodes, suggesting that the two node types are distinct in the learning process. Fig. \ref{fig:IGDistribution} shows the distribution of IG values. There was a significant difference in the distribution of IG values between disease and gene nodes. These results suggest that the node types are distinct in the training process.\\

Next, regarding the explainability of GraphIX in drug repositioning, we hypothesized that proteins for which the model gave a high contribution to the prediction of disease-drug associations might also be important in understanding actual pharmacological effects. We then quantitatively tested the validity of high-contribution proteins using OncoKB. Since the node types were found to be distinct, we focused here on the contribution of the protein nodes to assess the validity of explainability. Specifically, we examined how much the protein node with the highest IG value matched the known target protein in the disease-drug-target protein evaluation set. Table \ref{table:OncoKB} shows the OncoKB evaluation set, the number of candidate protein nodes used by the model for prediction, and the ranking of the IG value of the target proteins among the candidate proteins. The protein with the highest IG value matched the known target protein in 16 of the 21 sets. In other words, with 76\% accuracy, the protein nodes that contributed the most to the disease-drug association matched the known target proteins. This high correct answer rate indicates that GraphIX is able to capture the protein that are important for drug efficacy from the graphical structural information and that it has reasonable explainability in predicting drug repositioning.\\

\subsection{Case study}
\subsubsection{Disease-drug association prediction}
\begingroup
\renewcommand{\arraystretch}{1.2}
\begin{table}[h]
    \centering
    \begin{tabular}{cccc}
        \toprule
        Disease & Rank & Drug & evidence \\
        \midrule
        \multirow{5}{*}{Intestinal Neoplasms} 
        & 1 & Salicylic acid & Hamoya et al.(2016)\cite{Hamoya2016} \\
        & 2 & Leflunomide & Yamaguchi et al.(2019)\cite{Yamaguchi2019} \\
        & 3 & Quinacrine & Winer et al.(2021)\cite{Gallant2011} \\
        & 4 & Pentosan polysulfate & Enzyme et al.(1996)\cite{Anees1996} \\
        & 8 & Anakinra & Isambert et al.(2018)\cite{Isambert2018} \\
        \bottomrule
    \end{tabular}
    \caption{Candidate drugs with evidence for Intestinal Neoplasms.}
    \label{table:Disease-Drug}
\end{table}
\endgroup

For disease-drug novel association prediction, we describe a case study focusing on intestinal neoplasms. According to Global Cancer Statistics 2020\cite{Sung2021}, intestinal neoplasms are the second leading cause of cancer death worldwide and increase with socioeconomic development. \\
After excluding associations with drugs classified as dietary supplements and synonymous associations, the associations with intestinal neoplasms not included in the training data were ranked in descending order of association score. Then, we obtained the top 10 intestinal neoplasms - drug associations. 
As shown in Table \ref{table:Disease-Drug}, five of the top 10 associations were confirmed with the evidence from the literature (See Supplementary Table \ref{table:Disease-Drug-top10} for the top 10 novel associations with intestinal neoplasms).\\
Salicylic acid was a top-predicted candidate for intestinal neoplasms. It has been suggested that Nonsteroidal anti-inflammatory drugs (NSAIDs), including salicylic acid, may prevent colon cancer\cite{Hamoya2016}. Leflunomide was a second-ranked candidate drug. Yamaguchi et al. demonstrated that leflunomide inhibits the formation and growth of metastatic colonies of human-derived colorectal cancer tumors xenografted in mice\cite{Yamaguchi2019}. For the third-ranked candidate Quinacrine, Gallant et al. showed that Quinacrine reduced tumor deposition in nude mice with xenografts of human colorectal cancer\cite{Gallant2011}. Furthermore, a phase I study of quinacrine and capecitabine was conducted in patients with refractory metastatic colorectal cancer\cite{Winer2021}. In addition, GraphIX predicted that Pentosan polysulfate and Anakinra were associated with intestinal neoplasms, which was supported by the literature\cite{Anees1996}\cite{Isambert2018}. \\
These predicted drugs were not associated with cancer-indicating diseases in the training data (See Supplementary Table \ref{table:Disease-Drug-train} for the association details with these drugs in the training data). Therefore, the results show that GraphIX enables the prediction of new indication candidates completely different from the training data.\\

\begin{figure}[h]
\centering
\includegraphics[width=10cm]{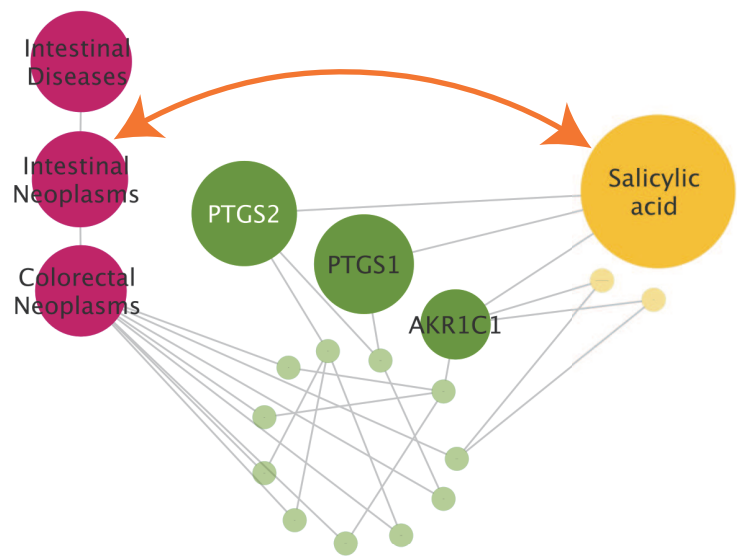}
\caption{Subgraph around the Intestinal Neoplasms-Salicylic acid association. Disease nodes, drug nodes, and gene nodes are depicted as red, yellow, and green circles, respectively. The circles with the name represent the nodes within 1-hop from the novel edge (Intestinal Neoplasms - Salicylic acid) and their size corresponds to the IG value on each node. The gene node with the highest IG value for the novel disease-drug association score is highlighted using white text.}
\label{fig:IntestinalNeoplasms_SalicylicAcid}
\end{figure}

Moreover, we investigated the explainability of the top predicted novel association "intestinal neoplasms - salicylic acid association". A subgraph connecting the Intestinal Neoplasms node and the Salicylic acid node with 5-hop simple paths is shown in Fig. \ref{fig:IntestinalNeoplasms_SalicylicAcid}. 
Prostaglandin-Endoperoxide Synthase 2(PTGS2) had the highest IG value for the Intestinal Neoplasms-Salicylic acid association score among the protein nodes. It is known that cancer cells overexpress PTGS2 in surrounding endothelial cells, leading to the promotion of angiogenesis and cancer growth, and it has been reported that the addition of salicylic acid to endothelial cells co-cultured with colon cancer cells decreases PTGS2 expression and inhibits endothelial tube formation\cite{Shtivelband2003}.
This case study confirmed that GraphIX could predict novel disease-drug associations and present target proteins important for drug efficacy as the most contributing entities.\\

\subsubsection{Gene-drug association prediction}
\begingroup
\renewcommand{\arraystretch}{1.2}
\begin{table}[h]
    \centering
    \begin{tabular}{cccc}
        \toprule
        Gene & Rank & Drug & evidence \\
        \midrule
        \multirow{4}{*}{IKBKB} 
        &                 1 & Leflunomide & Manna et al.(2000)\cite{Manna2000} \\
        &                 2 & Aminosalicylic acid & Yan et al.(1999)\cite{Yan1999} \\
        &                 4 & Etanercept & Das et al.(2012)\cite{Das2012} \\
        &                 7 & Bortezomib & Hideshima et al.(2009)\cite{Hideshima2009}\\
        \bottomrule
    \end{tabular}
    \caption{Candidate drugs with evidence for IKBKB.}
    \label{table:Gene-Drug}
\end{table}
\endgroup

For the gene-drug novel association case study, we focused on inhibitor of nuclear factor kappa B Kinase subunit beta (IKBKB), an important gene involved in inflammatory response propagation.
As in the disease-drug case study, associations with drugs classified as supplements were excluded, and gene-drug associations with IKBKB not included in the training data were ranked in descending order of predictive score to obtain the top 10 rankings. 
As shown in Table \ref{table:Gene-Drug}, four of the top 10 associations were confirmed with the evidence from the literature (See Supplementary Table \ref{table:Gene-Drug-top10} for top 10 novel predicted associations with IKBKB gene).\\
Leflunomide was the first-ranked candidate drug for the IKBKB gene. Leflunomide was the leading candidate for intestinal neoplasia. Manna et al.\cite{Manna2000} showed that Leflunomide inhibits activation of IKBKB by Tumor Necrosis Factor(TNF), confirming that this relationship has evidence in their paper. Aminosalicylic acid was the second-ranked candidate drug. Yan et al. indicated that Aminosalicylic acid inhibits the phosphorylation of IKBKB in mouse inflammatory bowel disease model cells\cite{Yan1999}. For the third-ranked candidate drug Etanercept, Das et al. demonstrated that Etanercept significantly reduced IKBKB activity in inflammatory nephropathy mice.\cite{Das2012}. In addition, the association between Bortezomib and IKBKB was supported by the literature\cite{Hideshima2009}\cite{Hideshima2014}.\\

\begin{figure}[!h]
\centering
\includegraphics[width=16cm]{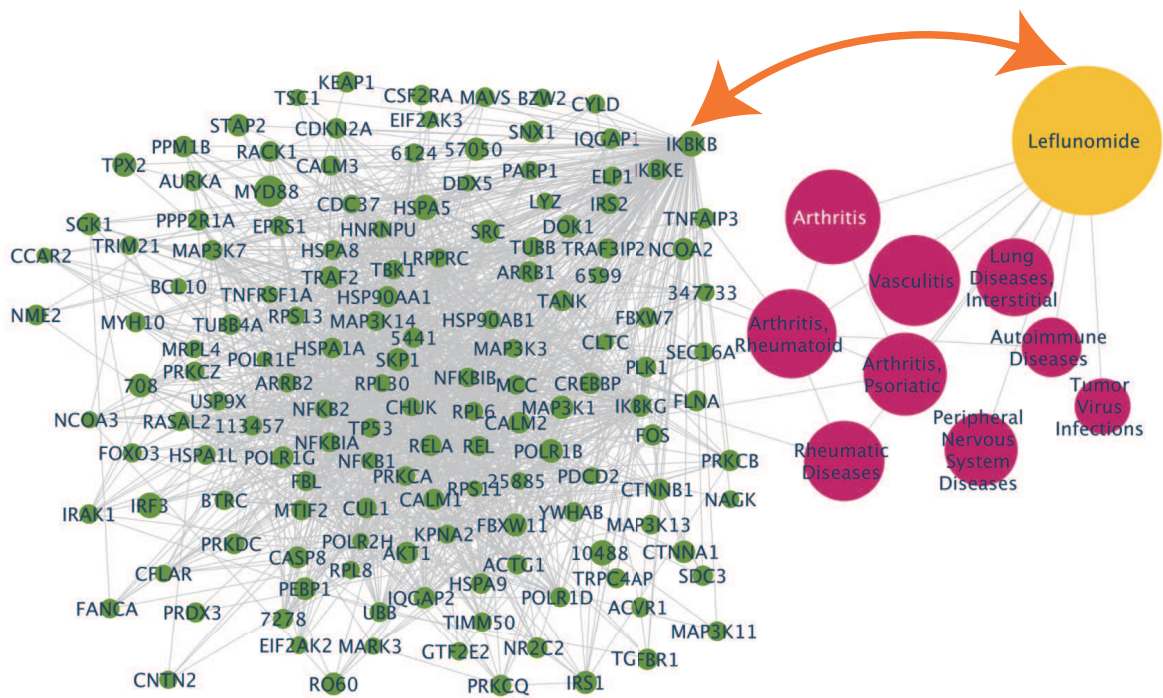}
\caption{Subgraph around IKBKB-Leflunomided novel edge. Disease nodes, drug nodes and gene nodes are depicted as red, yellow, and green circles, respectively. Circle size corresponds to the IG value on each node. The disease node with the highest IG value for the novel drug-gene association score is highlighted using whilte text.}
\label{fig:IKBKB_Leflunomide}
\end{figure}

The first-ranked novel association, IKBKB-Leflunomide, was validated for explainability utilizing IG values. A subgraph within 1-hop from IKBKB or Leflunomide is shown in Fig.\ref{fig:IKBKB_Leflunomide}. As shown in Fig.\ref{fig:IKBKB_Leflunomide}, Arthritis had the highest IG value for the IKBKB-Leflunomide association score among the disease nodes. Leflunomide is approved for the treatment of rheumatoid arthritis\cite{Smolen1999}, and was showed to provide a molecular basis for anti-inflammation through inhibition of IKBKB activation\cite{Manna2000}. In short, IKBKB and leflunomide are reasonable via arthritis. These results indicate that the GraphIX has reasonable novel association prediction and explainability in predicting gene-disease associations.\\

\section{Conclusion}
Since evidence-based drug discovery has become mainstream, understanding pharmacological actions cannot be cut in drug repositioning and the explainability of the prediction results may support appropriate insights in subsequent in-depth studies. Recently, in silico drug repositioning frameworks that take explainability into account have been proposed. This paper takes previous studies a step further by conducting a quantitative evaluation and validating the explainability in drug repositioning applications. Although this has not been considered in previous papers, confidence in XAI is an important factor while the validity of the explainability method is still under debate.\\

In the proposed framework GraphIX, disease-drug-gene relationships are learned on a biopharmaceutical knowledge graph. It then predicts the association between disease and drug and calculates the contribution values of proteins located in the neighborhood of the predicted disease-drug. Benchmark results show that GraphIX is comparable to recent supervised learning models. Quantitative validation of protein contribution values was performed using a real-world database including molecularly targeted drugs (drugs), cancer types (diseases), and target proteins (proteins). On average, 76\% of the proteins with the highest contribution values among the 14 candidate neighborhood proteins matched the actual drug target proteins for each disease and drug. The results indicate that GraphIX may be able to capture proteins of medicinal importance from biopharmacological knowledge graph structural information and that the explainability of the model is biologically reasonable. In the case studies, the predicted novel associations and their contributing entities were confirmed by the evidence from the literature.\\

There are three future tasks to be undertaken to further improve the utility of the model. The first is that by having the edges learn and predict the type of drug action on the disease, it may be possible not only whether there is an association, but also to distinguish whether it is a desirable therapeutic effect or a side effect to watch out for. Second is the reproducibility of predictions. Reproducibility of predictions is strongly required for AI in the life science field. GraphIX uses the TensorFlow\cite{Abadi2016} library, which is known to produce minute differences in numerical values during parallel computation on GPUs. While GraphIX prediction can be reproduced by specifying parameters, the reproducibility of AI is important in identifying novel associations that should proceed to experimental validation in the laboratory. Third, since the experimental discovery of protein interactions is advancing daily, more expanded PPIs based on the latest databases and literature will lead GraphIX to a more accurate presentation of genes contributing to the prediction results.\\

\section*{Code availability}
The open-source Python code to run the demo with GraphIX is available on GitHub\seqsplit{(https://github.com/clinfo/GraphIX.git)}.

\section*{CRediT authorship contribution statement}
Atsuko Takagi: Methodology, Software, Formal analysis, Investigation, Writing - original draft, Visualization. Mayumi Kamada: Investigation, Writing - review \& editing. Eri Hamatani: Methodology, Investigation. Ryosuke Kojima: Methodology, Software, Writing - review \& editing. Yasushi Okuno: Conceptualization, Writing - review \& editing.

\section*{Declaration of Competing Interest}
The authors declare that they have no known competing financial interests or personal relationships that could have appeared to influence the work reported in this paper.

\bibliographystyle{junsrt}
\bibliographystyle{elsarticle-harv}
\bibliography{ms}

\end{document}


\hrulefill
\vskip \baselineskip
\centering
\Large SUPPLEMENTARY MATERIAL OF\\
 GRAPHIX: GRAPH-BASED IN SILICO XAI FOR DRUG REPOSITIONING FROM BIOPHARMACEUTICAL NETWORK\\
\hspace*{12pt} \hrulefill

\vskip \baselineskip
\vskip \baselineskip
\begin{figure}[h]
\centering
\includegraphics[width=16cm]{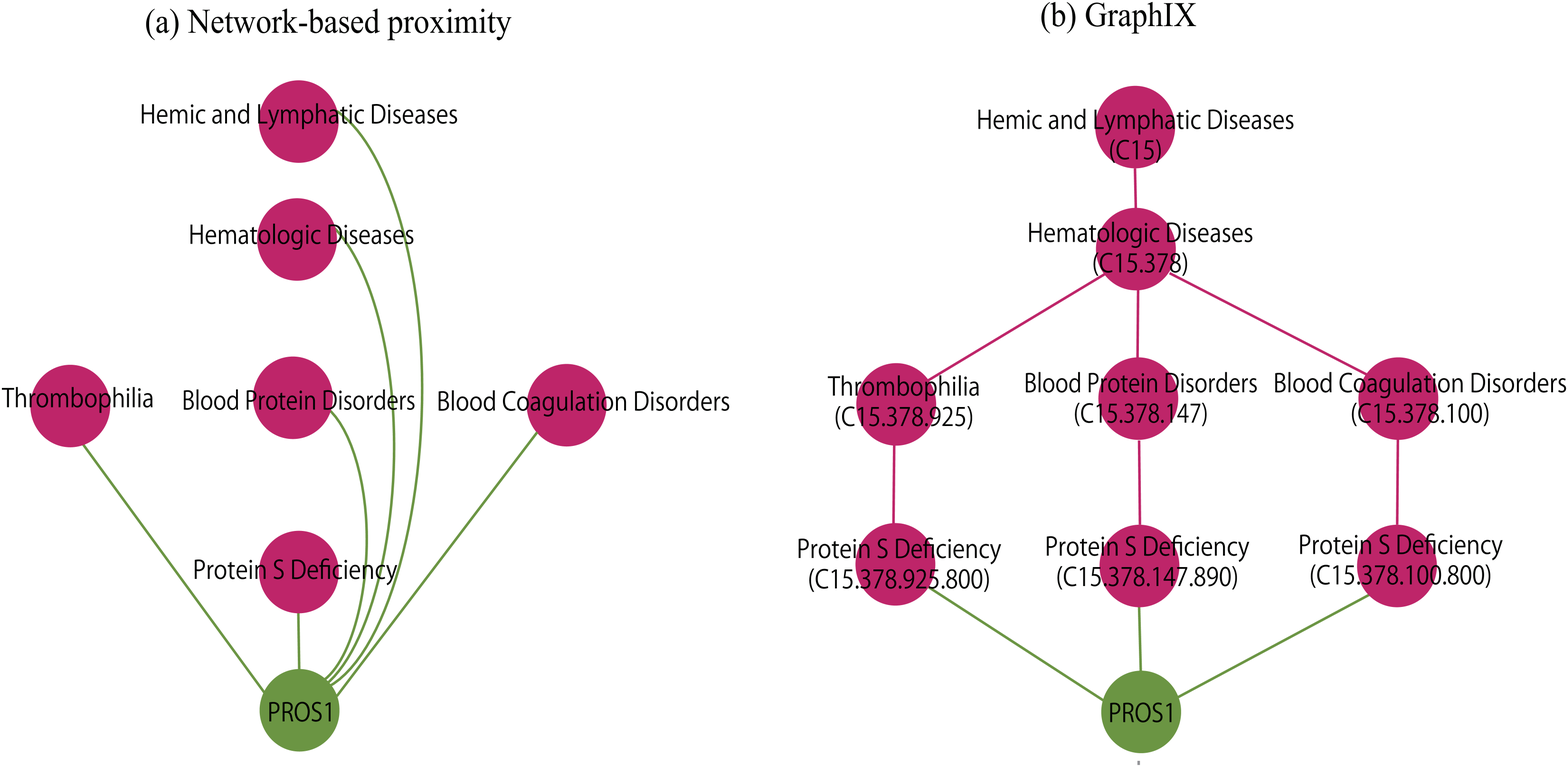}
\captionsetup{justification=raggedright,singlelinecheck=false}
\caption{Disease-gene association in (a) Network-based proximity and (b) GraphIX. Disease nodes and gene nodes are dipicted as red and green circles, respectively. Green lines and red lines indicate disease-gene and disease-disease assoiciations , respectively. Starting from (a), the disease names were converted to MeSH tree numbers. Then only disease-gene edges connecting to the most specific disease nodes were left by deleting obvious edges upward of them.  Finally, MeSH disease tree structure was added and (b) was obtained.}
\label{fig:KGwithMeSH}
\end{figure}

\clearpage
\begingroup
\renewcommand{\arraystretch}{1.4}
\begin{table}[h]
\label{table:Property}
\begin{minipage}{16cm}
\centering
\subcaption{Disease-drug prediction.}
\begin{tabular}{ccc}
\toprule
average shortest path length & average clustering coefficient & average betweeness centrality \\
\midrule
4.38 & $1.05 \times 10^{-1}$ & $1.38  \times 10^{-4}$ \\
\bottomrule
\end{tabular}
\end{minipage}

\begin{minipage}{16cm}
\vspace{0.5cm}
\end{minipage}

\begin{minipage}{16cm}
\centering
\subcaption{Drug-gene prediction.}
\begin{tabular}{ccc}
\toprule
average shortest path length & average clustering coefficient & average betweeness centrality \\
\midrule
3.94 & $1.25 \times 10^{-1}$ & $1.47  \times 10^{-4}$ \\
\bottomrule
\end{tabular}
\end{minipage}

\begin{minipage}{16cm}
\vspace{0.5cm}
\end{minipage}

\begin{minipage}{16cm}
\centering
\subcaption{Disease-gene prediction.}
\begin{tabular}{ccc}
\toprule
average shortest path length & average clustering coefficient & average betweeness centrality \\
\midrule
4.45 & $1.22 \times 10^{-1}$ & $1.43  \times 10^{-4}$ \\
\bottomrule
\end{tabular}
\end{minipage}
\caption{Network property.}
\end{table}
\endgroup

\clearpage
\begingroup
\renewcommand{\arraystretch}{1.0}
\begin{table}
    \centering
    \begin{tabular}{cll}
        \toprule
        Rank & Drug(New edge) & Disease(Training data) \\
        \midrule
        \multirow{7}{*}{1} & \multirow{7}{*}{Salicylic acid} & diarrhea \\
        & & keratoderma palmoplantar \\
        & & keratosis \\
        & & pain \\
        & & psoriasis \\
        & & skin diseases \\
        & & warts \\ \\
        \multirow{9}{*}{2} & \multirow{9}{*}{Leflunomide} & arthritis \\
        & & arthritis psoriatic \\
        & & arthritis arthritis rheumatoid \\
        & & autoimmune diseases \\
        & & lung diseases interstitial \\
        & & peripheral nervous system diseases \\
        & & rheumatic diseases \\
        & & tumor virus infections \\
        & & vasculitis \\ \\
        \multirow{11}{*}{3} & \multirow{11}{*}{Quinacrine} & abnormalities multiple \\
        & & chromosome aberrations \\
        & & chromosome disorders \\
        & & creutzfeldt-jakob syndrome \\
        & & intellectual disability \\
        & & lupus erythematosus systemic \\
        & & malaria \\
        & & pemphigus \\
        & & pneumothorax \\
        & & prion diseases \\
        & & turner syndrome \\ \\        
        \multirow{7}{*}{4} & \multirow{7}{*}{Pentosan polysulfate} & creutzfeldt-jakob syndrome \\
        & & osteoarthritis \\
        & & pain \\
        & & prion diseases \\
        & & thrombocytopenia \\
        & & thrombosis \\
        & & urinary calculi \\ \\
        \multirow{4}{*}{8} & \multirow{4}{*}{Anakinra} & arthritis rheumatoid \\
        & & cryopyrin-associated periodic syndromes \\
        & & familial mediterranean fever \\
        & & mevalonate kinase deficiency \\
        \bottomrule
    \end{tabular}
    \caption{Drug(new edge) shows candidate drugs for Intestinal Neoplasms, and Disease(training data) shows diseases related to the drug in the training data. }
    \label{table:Disease-Drug-train}
\end{table}
\endgroup

\clearpage
\begingroup
\renewcommand{\arraystretch}{1.2}
\begin{table}[h]
    \centering
    \begin{tabular}{cccc}
        \toprule
        Disease & Rank & Score & Drug \\
        \midrule
        \multirow{10}{*}{\begin{tabular}{c}Intestinal \\Neoplasms\end{tabular}} 
        & 1 & 7.37 & Salicylic acid \\
        & 2 & 7.11 & Leflunomide \\
        & 3 & 7.10 & Quinacrine \\
        & 4 & 6.89 & Pentosan polysulfate \\
        & 5 & 6.86 & Clobetasol propionate \\
        & 6 & 6.75 & Oseltamivir \\
        & 7 & 6.63 & Epoprostenol \\
        & 8 & 6.59 & Anakinra \\
        & 9 & 6.48 & Hydrocodone \\
        & 10 & 6.45 & Pyrimethamine \\
        \bottomrule
    \end{tabular}
    \caption{Top 10 ranked drugs for Intestinal Neoplasms.}
    \label{table:Disease-Drug-top10}
\end{table}
\endgroup

\begingroup
\renewcommand{\arraystretch}{1.2}
\begin{table}[h]
    \centering
    \begin{tabular}{cccc}
        \toprule
        \quad Gene\quad\quad & Rank\quad\quad & \quad Score\quad & \quad Drug\quad \\
        \midrule
        \multirow{10}{*}{IKBKB} 
        & 1 & 14.35 & Leflunomide \\
        & 2 & 14.30 & Aminosalicylic acid \\
        & 3 & 13.89 & Lenalidomide \\
        & 4 & 13.75 & Etanercept \\
        & 5 & 13.19 & Pomalidomide \\
        & 6 & 13.07 & Adalimumab \\
        & 7 & 12.95 & Bortezomib\\
        & 8 & 12.58 & Rituximab \\
        & 9 & 12.50 & Ibuprofen \\
        & 10 & 11.99 & Hyaluronic acid \\
        \bottomrule
    \end{tabular}
    \caption{Top 10 ranked drugs for IKBKB.}
    \label{table:Gene-Drug-top10}
\end{table}
\endgroup

\clearpage
\begin{figure}[h]
\centering
\includegraphics[width=12cm]{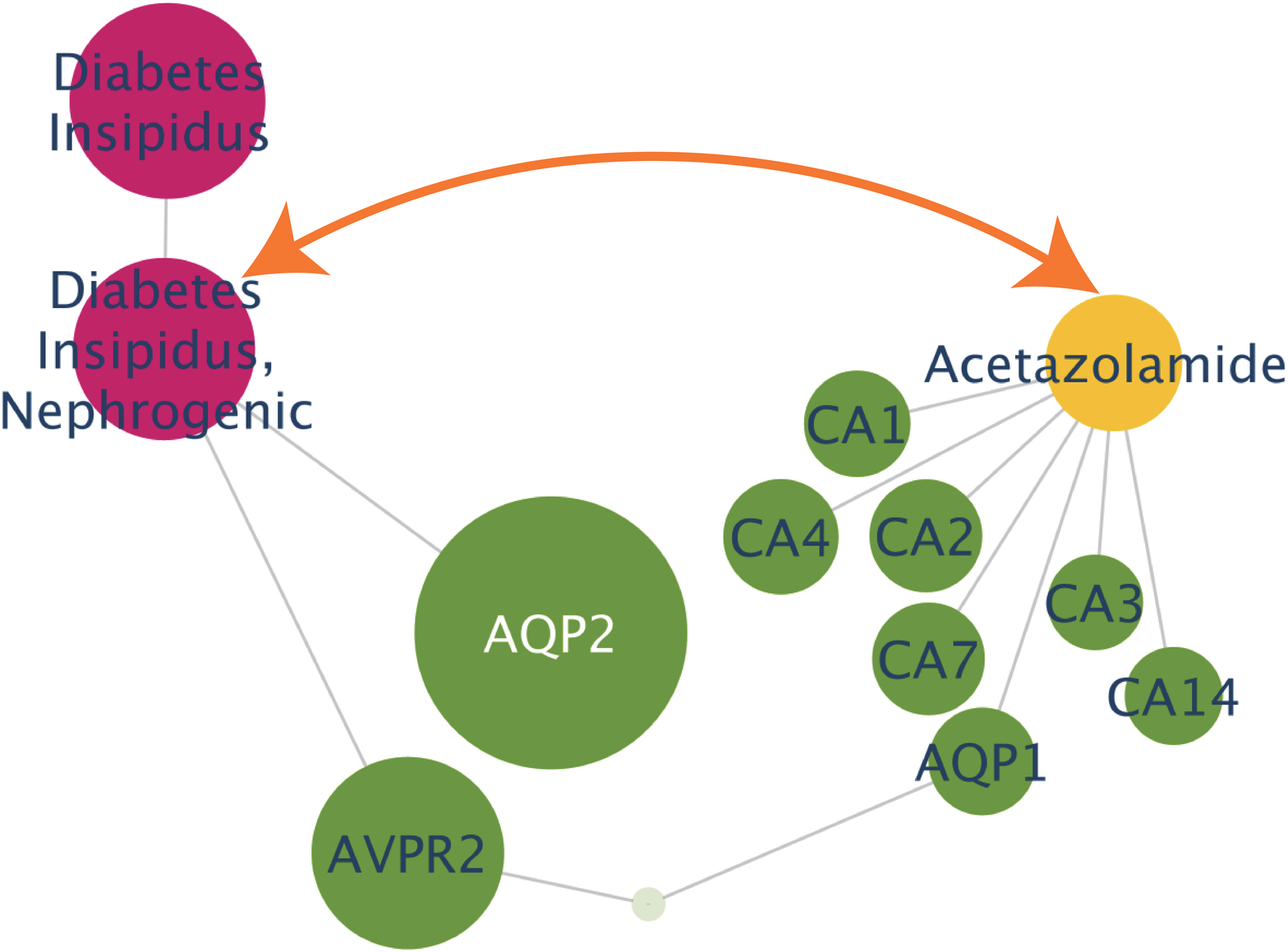}
\captionsetup{justification=raggedright,singlelinecheck=false}
\caption{Subgraph around Nephrogenic Diabetes Insipidus(NDI)-Acetazolamide novel edge. Disease nodes, drug nodes and gene nodes are dipicted as red, yellow and green circles, respectively. The circles with the name represent the nodes within 1-hop from the new edge and their size corresponds to IG value on each node. The transparent circles represent a node outside 1-hop, where IG cannot be calculated. The gene node with the highest IG value for the novel disease-drug prediction score is highlighted using white text. AQP2 is the biomarker of Acedazolamide for NDI\cite{Groot2016}.}
\label{fig:NDI-Acetazolamide}
\end{figure}

\bibliographystyle{elsarticle-harv}
\bibliography{supplement}